\documentclass[conference]{IEEEtran}
\ifCLASSINFOpdf
\else
\fi

\usepackage{tikz}

\usepackage[flushleft]{threeparttable}
\usepackage{url}
\usepackage{amsmath}
\usepackage{multirow}
\usepackage{subcaption}
\usepackage{fancybox}
\usepackage{graphicx}
\usepackage{color}
\usepackage{xcolor}

\usepackage{caption,graphicx,newfloat}
\DeclareCaptionType{List}

\usepackage{graphicx,calc}
\newlength\myheight
\newlength\mydepth
\settototalheight\myheight{Xygp}
\begin{document}
%
% paper title
% Titles are generally capitalized except for words such as a, an, and, as,
% at, but, by, for, in, nor, of, on, or, the, to and up, which are usually
% not capitalized unless they are the first or last word of the title.
% Linebreaks \\ can be used within to get better formatting as desired.
% Do not put math or special symbols in the title.
\title{On the Resilience of RTL NN Accelerators:\\Fault Characterization and Mitigation}

\author{\IEEEauthorblockN{Behzad Salami\IEEEauthorrefmark{1}\IEEEauthorrefmark{2},
Osman Unsal\IEEEauthorrefmark{1}, and
Adrian Cristal\IEEEauthorrefmark{1}\IEEEauthorrefmark{2}}
\IEEEauthorblockA{\IEEEauthorrefmark{1}Barcelona Supercomputing Center (BSC), Barcelona, Spain. Emails: \{behzad.salami, osman.unsal, and adrian.cristal\}@bsc.es}
\IEEEauthorblockA{\IEEEauthorrefmark{2}Universitat Politècnica de Catalunya (UPC), Barcelona, Spain.}
}

\maketitle

% As a general rule, do not put math, special symbols or citations
% in the abstract
%\begin{abstract}
%The abstract goes here.
%\end{abstract}

\begin{abstract}
Machine Learning (ML) is making a strong resurgence in tune with the massive generation of unstructured data which in turn requires massive computational resources. Due to the inherently compute- and power-intensive structure of Neural Networks (NNs), hardware accelerators emerge as a promising solution. However, with technology node scaling below 10nm, hardware accelerators become more susceptible to faults, which in turn can impact the NN accuracy. In this paper, we study the resilience aspects of Register-Transfer Level (RTL) model of NN accelerators, in particular fault characterization and mitigation. By following a High Level Synthesis (HLS) approach,  \textit{first}, we characterize the vulnerability of various components of RTL NN. We observed that the severity of faults depends on both \textit{i}) application-level specifications, i.e., NN data (inputs, weights, or intermediate), NN layers, and NN activation functions, and \textit{ii}) architectural-level specifications, i.e., data representation model and the parallelism degree of the underlying accelerator. \textit{Second}, motivated by characterization results, we present a low-overhead fault mitigation technique that can efficiently correct bit flips, by 47.3\% better than state-of-the-art methods.
\end{abstract}
\section{Introduction}
\label{sec:intro}
%\footnote{\textit{\textbf{New Paper, Not an Extension of a Conference Paper}}}
Machine learning models and in particular Neural Networks (NNs) are increasingly being used in the context of nonlinear "cognitive" problems, such as natural language processing and computer vision. These models can learn from a dataset in the training phase and make predictions on a new, previously unseen data in the inference/prediction/classification phase with an ever-increasing accuracy. %Hence, these models need to go through a training phase, during which their parameters (weights and biases) are adjusted before they are able to classify or recognize previously unseen inputs. {\color{black}There are diverse models of neural networks targeted for various contexts, e.g., Recursive- (RNN), Convolutional- (CNN), and fully-connected Neural Networks, among other models \cite{sze2017efficient}. They are distinguished by distinct neural intra- and inter-connection structure and computational models, which lead to various capabilities in the feature extraction and transformation. In this paper, we focus on the fully connected model, since it is a common type, accurate enough for mid-scale classifications, widely used in the structure of other models like CNNs, and has a straightforward computational model.\footnote{For the rest of the paper, we will refer to fully connected neural network model as DNN.}. 
%Typically, NNs encompass two distinct phases: training and inference (prediction/classification). Training is basically a one-time process to determine parameters of NN. In contrast, the classification phase runs repeatedly to infer unknown data.
However, the compute- and power-intensive nature of NNs prevents their effective deployment in resource-constrained environments, such as mobile scenarios \cite{mobile}. Hardware acceleration on Application Specific Integrated Circuits (ASICs) or Field Programmable Gate Arrays (FPGAs) offers a roadmap for enabling NNs in these scenarios \cite{suda2016throughput}, \cite{li2016high}, \cite{minerva}, \cite{eyeriss}, \cite{yodann}. However, similar to general purpose devices, hardware accelerators are also susceptible to faults (permanent/hard and transient/soft), as a consequence of Single Event Upset (SEU), manufacturing defects, and below safe-voltage operations. The ever-increasing rate of these faults especially in nano-scale technology nodes can directly impact the accuracy of NNs. 

Traditionally, to perform early studies and apply further optimizations, application-specific hardware designs are modeled in different abstraction levels before the final in-silicon implementation. For instance, a hardware design can be modeled, e.g,. in software-level behavioral simulator, functional level, Transaction-Level Model (TLM), Register-Transfer Level (RTL), and transistor-level. Among them, thanks to the evolution of High-Level Synthesis (HLS) tools to abstract low-level complexities of the hardware \cite{nane2016survey}, the RTL model has received significant attention. This approach can lead to a decreased development time with an early evaluation of the final design, while conforming to final power, energy, performance, and resilience goals in comparison to the in-silicon ASIC/FPGA implementation. For instance, an HLS-based approach has been used in recent research to study the resilience of accelerators \cite{tambara2017}, \cite{lee2017}, \cite{tonfat2017}, \cite{dossantos2017}. In this paper, we also use an HLS-based approach to study accelerator resilience. However we use the HLS approach to study the fault characterization and mitigation of RTL model of NNs. Understanding this resilience behavior can provide an opportunity for the further resilience studies on the in-silicon NN accelerators. 
Main contributions of this work are summarized as follows:  
\begin{itemize}
%\item A fault injection framework on top of our parameterized RTL design of a DNN classifier, with support for permanent and transient faults.    
\item Fault Characterization: We perform an in-depth vulnerability study in the various components of the RTL NN accelerators against permanent and transient faults. Our experiments indicate that both application-level specifications, i.e., NN data (inputs, weights, or intermediate), NN layers, and NN activation functions, as well as architectural-level specifications, i.e., data representation model and parallelism degree of the underlying accelerator, have significant impact on the severity of faults.
%\begin{itemize}
%\item The sign-bit is 4.1X by on average more vulnerable than the digit and fraction components when low-precision fixed-point data representation model is used.\footnote{The component A of the RTL NN is N times more vulnerable than component B means that faults injected in the component A causes on average N times higher classification errors, in comparison to the same number of faults injected in the component B.}.
%\item Weights are 2.9X by on average more vulnerable than input data.
%\item In the presence of faults, exploiting logarithmic sigmoid as the activation function leads to by on average XX less classification error than ReLu.
%\item In the presence of transient faults, inner layers (closer to the output layer of the NN) are relatively more vulnerable.
%\end{itemize}
\item Fault Mitigation: Motivated by the fault characterization experimental results, we present a low-overhead technique to mitigate faults by recovering corrupted bits, without any need for redundant data. The efficiency of the proposed technique is by 47.3\% better than the state-of-the-art methods. 
\end{itemize} 

The rest of the paper is organized as follows. In Section II the overall methodology, i.e., the RTL NN and fault model, is introduced. %We elaborate the architecture of the RTL design of the DNN model and the fault injection unit in Section \ref{sec:model} and illustrate them in Section \ref{sec:examples}. The baseline configuration and the analysis of the fault propagation is explained in Section \ref{sec:baseline}. 
The fault characterization and mitigation studies are discussed in Section III and Section IV, respectively. %we introduce and evaluate our fault mitigation techniques. 
We review the previous work in Section V and finally, the paper is summarized and concluded in Section VI. 
\section{Introducing the Overall Methodology}
\label{sec:overall}
This section presents our methodology to conduct the resilience study, i.e., the architecture of RTL NN accelerator and fault model.

%figure
\begin{figure*}
\centering
\includegraphics[width=\textwidth]{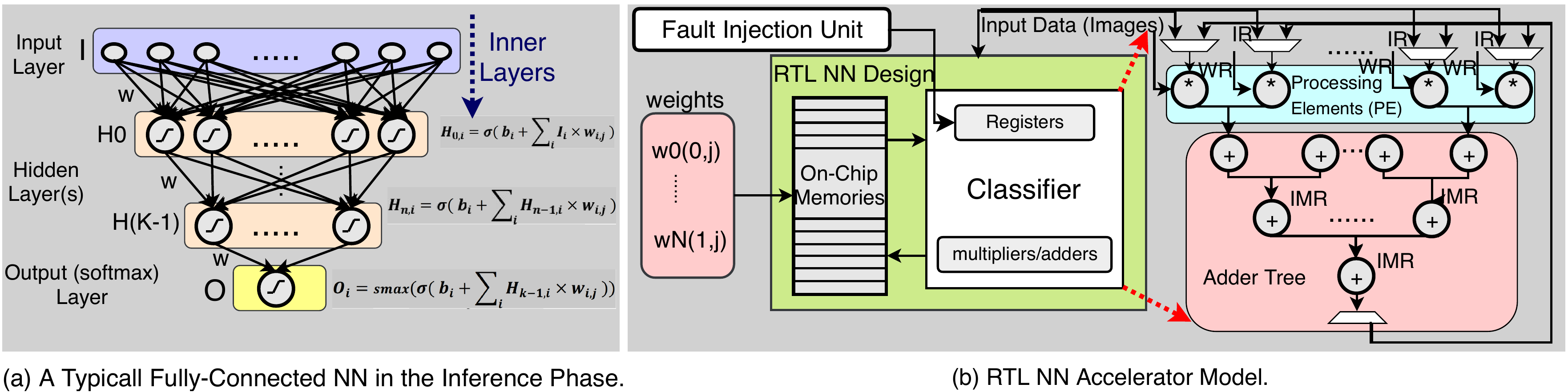} %, height=5cm
\caption{{\color{black}The Overall Methodology to Resilience Study of the RTL NN Accelerator. }}
\label{fig:overall}
\end{figure*}

\subsection{The Architecture of RTL NN Accelerator}
Specifications of the experimented RTL NN with a baseline configuration is summarized in Table \ref{table:nnsetup}. Our study features a typical fully-connected NN that is also widely used in the structure of other NN models \cite{minerva}. Our study targets the inference phase of NN since training is normally a one-time process; additionally, the inference is repeatedly performed to classify unknown data. As can be seen in Fig. \ref{fig:overall}(a), this NN model is composed of input, hidden, and output layers, where all adjacent layers are fully connected to each other. The first/last layer is the input/output layer and has one neuron for each component in the input/output vector. Between the input and output layers, there are single/multiple hidden layers. The interconnection between neurons of adjacent layers is determined based on a collection of weights and biases, whose values are tuned in the training phase. Each NN neuron uses an activation function to determine its output. Finally, in the output layer, a softmax function generates the final output of the NN. We perform our experiments on a 6-layer NN, i.e., ($\{L_i, i \in [0,5]\}$), one input, four hidden, and one output layer(s). The four hidden layer sizes are fixed at 1024, 512, 256, 128 while input and output layer sizes are benchmark-dependent (784, 54 and 2437 for input while 10, 8 and 52 for output layers for the three NN applications studied in this paper, i.e., MNIST \cite{mnist}, Forest \cite{forest}, and Reuters \cite{reuters}, respectively.). Thus, there are five matrix multipliers among adjacent layers, i.e., ($\{Layer_j, j \in [0,4]\}$), where $Layer_j$ refers to the matrix multiplication of $L_j$ and $L_{j+1}$. Among benchmarks, MNIST is a set of black and white digitized handwritten digits, each image composed of 784*8-bit pixels, the output infers the number from 0 to 9 (10 output classes), with 60000 training- and 10000 inference images. Forest includes cartographic observations for classifying the forest cover type. Reuters covers news articles for text categorization. MNIST is most widely-used by the ML community to evaluate the efficiency of novel NN methods. Hence, we use MNIST as the main benchmark to evaluate our resilience studies. To demonstrate the generality of experimental observations, we briefly present results for Forest and  Reuters, as well.
\begin{table}[]
\centering
\caption{Detailed Specifications of the Baseline RTL NN Setup.}
\label{table:nnsetup}
\begin{tabular}{|l|l|}
\hline
\multicolumn{2}{|c|}{\textbf{Neural Network (NN)}}                                                 \\ \hline \hline
\textbf{Type}                       & Fully-Connected Classifier                  \\ 
\textbf{Topology (number of layers)}        & 6L (1L input, 4L hidden, 1L output)                     \\ 
\textbf{Per Layer Size (number of neurons)} & (784, 1024, 512, 256, 128, 10)= 2714                    \\ 
\textbf{Total Number of Weights}                  & $\sim$1.5 million                                      \\ 
\textbf{Activation Function}        & Logarithmic Sigmoid (logsig)                                      \\ \hline \hline
\multicolumn{2}{|c|}{\textbf{Benchmark}}                                                      \\ \hline \hline
\textbf{Name-Type}                  & MNIST \cite{mnist}- Handwritten Digits \\ 
\textbf{Number of Images}                   & Training: 60000, Inference: 10000                  \\ 
\textbf{Number of Pixels per Image}         & 28*28= 784                                              \\ 
\textbf{Number of Output Classes}                  & 10                                                      \\ \hline \hline
\multicolumn{2}{|c|}{\textbf{Data Representation Model}}                                      \\ \hline \hline
\textbf{Type}                       & 16-bits Fixed-Point (FP)                                     \\ 
\textbf{Precision}                  & Min sign and digit per layer (Fig. \ref{fig:precision})              \\ \hline \hline
\multicolumn{2}{|c|}{\textbf{An Example Synthesize of RTL NN on FPGA}}                                            \\ \hline \hline
\textbf{FPGA Platform-Chip}              & VC707-Virtex7                                                   \\ 
\textbf{Operating Frequency}        & 100Mhz                                                  \\ 
\textbf{BRAM Usage} (Total: 2060)   & 70.8\%                                                  \\
\textbf{DSP Usage} (Total: 2800)    & 8.6\%                                                  \\ 
\textbf{FF Usage} (Total: 303,600)  & 3.8\%                                                  \\ 
\textbf{LUT Usage} (Total: 607,200) & 4.9\%                                                  \\ 
\textbf{Number of $PE$s} & 64
\\
\hline
\end{tabular}
\end{table}

We build the RTL NN leveraging Bluespec \cite{bluespec}, which is a state-of-the-art rule-based cycle-accurate HLS ecosystem including the BSV language together with the BSC compiler which can produce Verilog. As can be seen in Fig. \ref{fig:overall}(b), our RTL design is composed of \textit{i}) on-chip memories to accommodate weights, \textit{ii}) different set of registers to latch data during the inference, \textit{iii}) a set of multiplier-adder processing elements ($PE$s) and an adder-tree to perform the required matrix multiplications, and \textit{iv}) a piece-wise linear model of the activation function. This is a typical model of NN accelerators, as surveyed in \cite{sze} for ASICs and in \cite{guo} for FPGAs. Through this setup, input data is streamed through $PE$s in parallel to perform the matrix multiplications in a fully-pipelined fashion. The number of clock cycles to classify an input object item can be computed as a function of $|L_i|$ and the number of $PE$s, as shown in Eq. \ref{T}.\footnote{For instance, by exploiting 64 $PE$s in our baseline RTL NN, $T$ can be computed as; $T=\frac{784*1024+1024*512+512*256+256*128+128*10}{64}=1490944$.} 

\begin{equation}
\label{T}
T = \frac{\sum_{i=0}^{N_l-2} {|L_i| * |L_{i+1}|}}{\#PEs}
\end{equation}

It is important to note that our design is fully paramterisable on the number and size of NN layer and the number of $PEs$; also, it is fully synthesizable. For instance, the area utilization of synthesizing it on VC707, a Xilinx Virtex7 technology is shown in Table \ref{table:nnsetup}. Furthermore, among pipeline stages of the RTL design, various registers are exploited to latch different types of NN data. At each cycle, a different data item is latched to these registers. As can be seen in Fig. \ref{fig:overall}(b), there are three types of registers, i.e., input registers (\textit{IR}s), weight registers (\textit{WR}s), and intermediate registers (\textit{IMR}s), which are leveraged to latch input, weights, and intermediate data, respectively. %Also, a fault injection unit is developed to manage fault injection in registers of the RTL design, supporting permanent (stuck-at-0 and stuck-at-1) and transient faults.

\begin{figure}
  \includegraphics[width=0.9\linewidth]{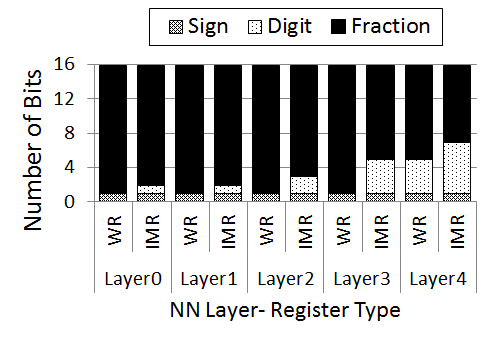}
  \caption{Minimum Precision to Represent Data of RTL NN, i.e., Inputs ($IR$s), Weighs ($WR$s), and Intermediate ($IMR$s). ($IR$s are composed of only fraction components since input data are in [0,1) range.)}
  \label{fig:precision}
\end{figure}

For experiments, we first export weights and biases of the trained NN that is performed off-line using a MATLAB implementation, initialize on-chip memories of the RTL design, and then start streaming 10000 input images to perform the inference. Also, for representing data, we use the fixed-point low-precision model. Note that lowering the precision of data is a common technique for applications in the approximate computing domain, in particular for NNs performing inference \cite{fpgafp}, to achieve power and performance efficiency with negligible accuracy loss. Following this approach, we use per-layer minimum precision fixed-point model. The bit-width of data (input, weights, and intermediate) is fixed to 16-bits, composed of the sign, digit, and fraction components. Toward this goal, with a pre-processing analysis, we extract the minimum bit-widths of the sign and digit components per layer and the rest of 16 bits are filled by the fraction component. As we experimentally observed, this quantification does not lead to any considerable accuracy loss in comparison to a full-precision data model. The minimum precision is used in this paper is summarized in Fig. \ref{fig:precision}. Also, by following the default internal approach of Bluespec FixedPoint (FP) library, negative numbers ($<0$) are represented in two's complement model. % except the last layer, $Layer_4$, weights of other layers are in (-1, 1) margin, which means that there is no need for the digit component. However, for the $Layer_4$, a 4-bit digit component is used, the minimum width to represent data without any NN accuracy loss. 

%The baseline adopted NN is composed of six layers, one input, four hidden, and one output layer(s). Their sizes are 784, 1024, 512, 128, 64, and 10, respectively. Thus, there are five matrix multipliers: $layer_0$ for multiplying the input and hidden layers and $layer_1$ for multiplying the hidden and output layers. The total number of clock cycles to classify an image can be calculated as $T = \frac{784*128 + 128*10}{64} = 1588$ cycles, as formulated in Equation \ref{T}. %Also, the default activation function is \textit{logsig}, which non-linearly bounds data in the [0, 1] range. With respect to the nonlinear structure of the \textit{logsig}, we use a piece-wise linear implementation in our HLS development. The output of the matrix multiplication in the last layer of the DNN, $layer_1$, is a vector of 10 numbers and the final output is generated after a softmax function is applied on them. At the end, the classification error is computed as the percentage of missclassified images. 

\subsection{Fault Model}
As can be seen in Fig. \ref{fig:overall}, the fault injection unit is developed on top of the RTL NN classifier to manage fault injection into registers, i.e., $IR$s, $WR$s, and $IMR$s. It is important to note that other components of the RTL NN, e.g,. on-chip memory and computing $PE$s are also susceptible to faults. However, many resilience solutions have been developed for these components, such as Error Correction Code (ECC) for memories \cite{ecc} and time-staggered Razor shadow latches for $PE$s \cite{Stott2014}; thus, we perform this study by fault injecting in registers, the main state-holding elements in the RTL design. 

\begin{figure}
\centering
\begin{subfigure}{0.24\textwidth}
\centering
\includegraphics[width=\linewidth]{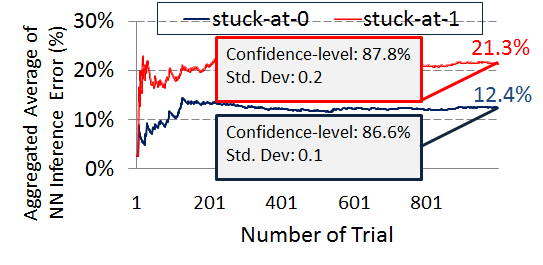} %, height=5cm
\caption{Permanent Faults.}
\label{fig:run_permanent}
\end{subfigure}
\hfill
\begin{subfigure}{0.24\textwidth}
\centering
\includegraphics[width=\linewidth]{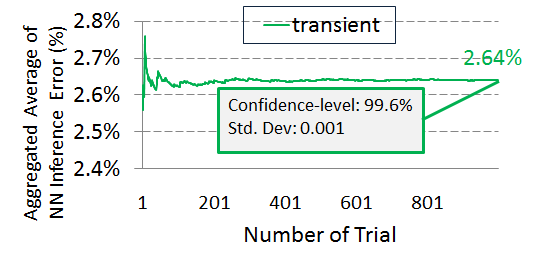} %, height=5cm
\caption{Transient Faults.}
\label{fig:run_transient}
\end{subfigure}
\caption{{\color{black}Repeating Trials to Achieve Statistically Significant Results. (Shows results of single-bit fault trials)}}
\label{fig:runs}
\end{figure}

\begin{figure*}
\centering
\begin{subfigure}{0.32\textwidth}
\centering
\includegraphics[width=\linewidth]{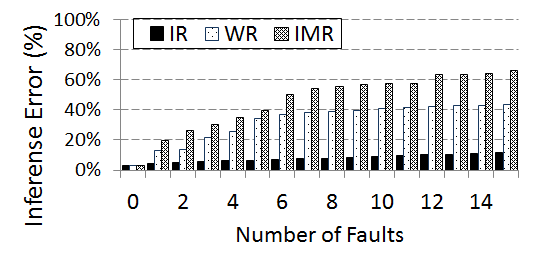} %, height=5cm
\caption{NN Data: stuck-at-0}
\label{fig:103}
\end{subfigure}
\begin{subfigure}{0.32\textwidth}
\centering
\includegraphics[width=\linewidth]{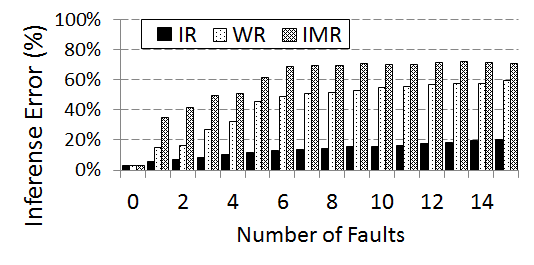} %, height=5cm
\caption{NN Data: stuck-at-1}
\label{fig:104}
\end{subfigure}
\begin{subfigure}{0.32\textwidth}
\centering
\includegraphics[width=\linewidth]{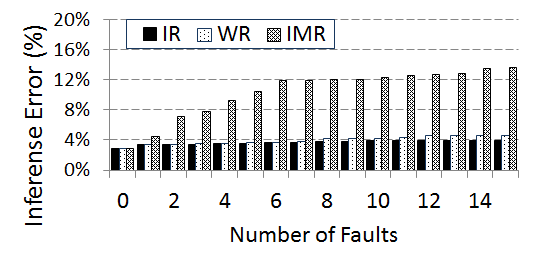} %, height=5cm
\caption{NN Data: transient}
\label{fig:105}
\end{subfigure}
\begin{subfigure}{0.32\textwidth}
\centering
\includegraphics[width=\linewidth]{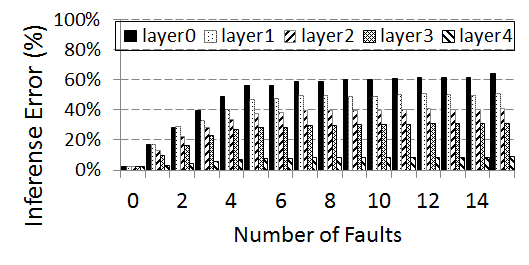} %, height=5cm
\caption{NN Layers: stuck-at-0}
\label{fig:106}
\end{subfigure}
\begin{subfigure}{0.32\textwidth}
\centering
\includegraphics[width=\linewidth]{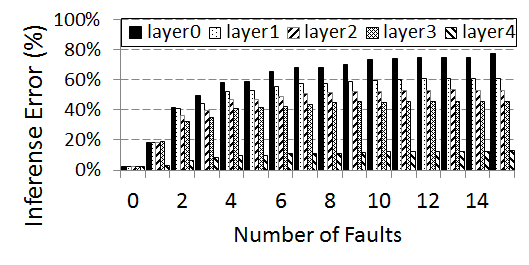} %, height=5cm
\caption{NN Layers: stuck-at-1}
\label{fig:107}
\end{subfigure}
\begin{subfigure}{0.32\textwidth}
\centering
\includegraphics[width=\linewidth]{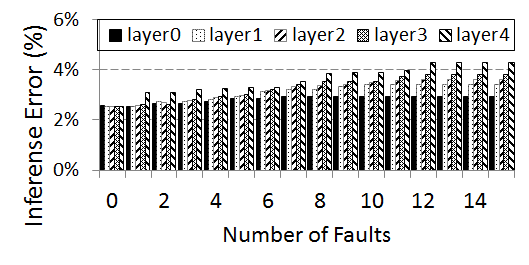} %, height=5cm
\caption{NN Layers: transient}
\label{fig:108}
\end{subfigure}
\begin{subfigure}{0.32\textwidth}
\centering
\includegraphics[width=\linewidth]{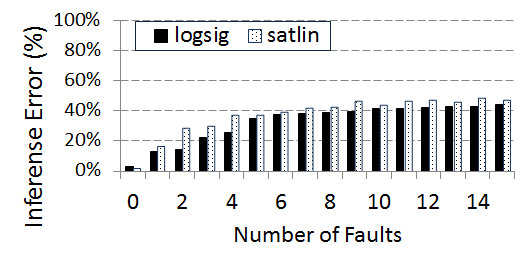} %, height=5cm
\caption{NN Functions: stuck-at-0}
\label{fig:112}
\end{subfigure}
\begin{subfigure}{0.32\textwidth}
\centering
\includegraphics[width=\linewidth]{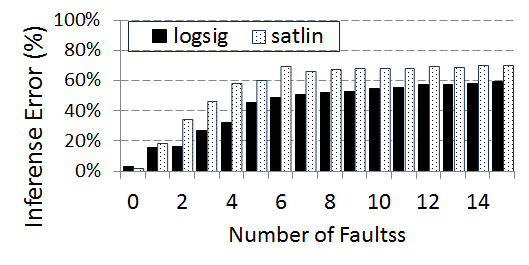} %, height=5cm
\caption{NN Functions: stuck-at-1}
\label{fig:113}
\end{subfigure}
\begin{subfigure}{0.32\textwidth}
\centering
\includegraphics[width=\linewidth]{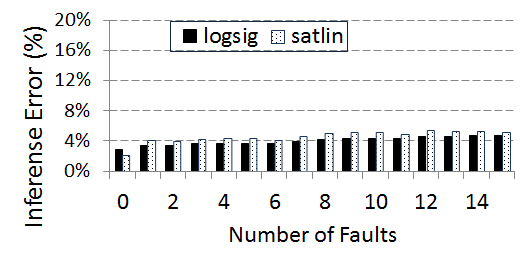} %, height=5cm
\caption{NN Functions: transient}
\label{fig:114}
\end{subfigure}
\begin{subfigure}{0.32\textwidth}
\centering
\includegraphics[width=\linewidth]{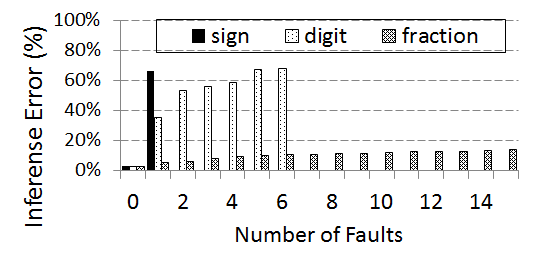} %, height=5cm
\caption{FP Components: stuck-at-0}
\label{fig:100}
\end{subfigure}
\begin{subfigure}{0.32\textwidth}
\centering
\includegraphics[width=\linewidth]{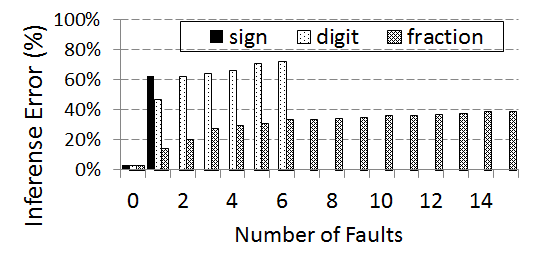} %, height=5cm
\caption{FP Components: stuck-at-1}
\label{fig:101}
\end{subfigure}
\begin{subfigure}{0.32\textwidth}
\centering
\includegraphics[width=\linewidth]{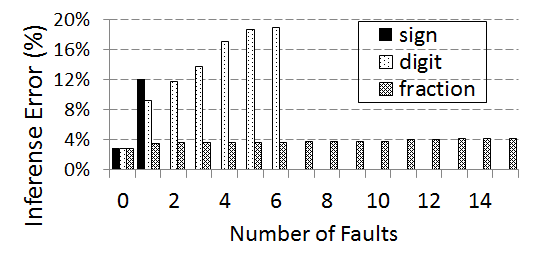} %, height=5cm
\caption{FP Components: transient}
\label{fig:102}
\end{subfigure}
\begin{subfigure}{0.32\textwidth}
\centering
\includegraphics[width=\linewidth]{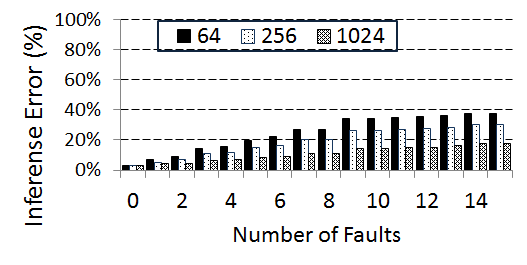} %, height=5cm
\caption{Number of $PE$s: stuck-at-0}
\label{fig:109}
\end{subfigure}
\begin{subfigure}{0.32\textwidth}
\centering
\includegraphics[width=\linewidth]{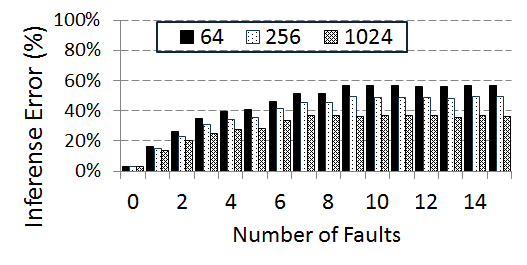} %, height=5cm
\caption{Number of $PE$s: stuck-at-1}
\label{fig:110}
\end{subfigure}
\begin{subfigure}{0.32\textwidth}
\centering
\includegraphics[width=\linewidth]{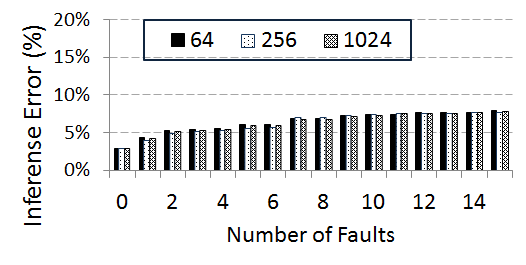} %, height=5cm
\caption{Number of $PE$s: transient}
\label{fig:111}
\end{subfigure}
\begin{subfigure}{0.32\textwidth}
\centering
\includegraphics[width=\linewidth]{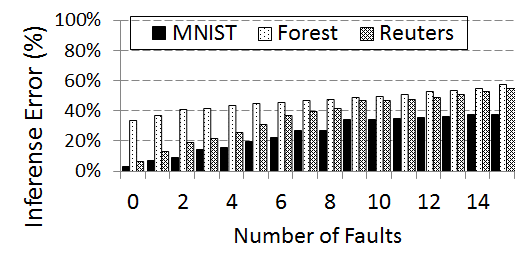} %, height=5cm
\caption{Datasets: stuck-at-0}
\label{fig:120}
\end{subfigure}
\begin{subfigure}{0.32\textwidth}
\centering
\includegraphics[width=\linewidth]{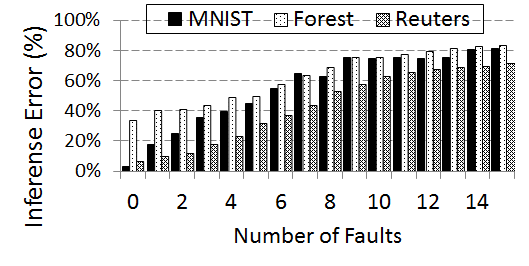} %, height=5cm
\caption{Datasets: stuck-at-1}
\label{fig:121}
\end{subfigure}
\begin{subfigure}{0.32\textwidth}
\centering
\includegraphics[width=\linewidth]{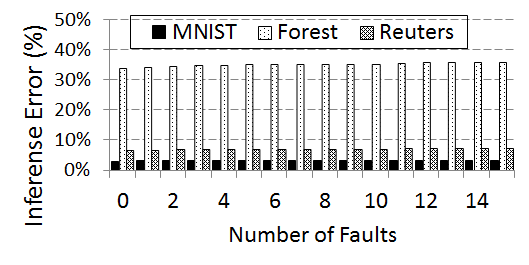} %, height=5cm
\caption{Datasets: transient}
\label{fig:122}
\end{subfigure}
\caption{{\color{black}Fault Characterization in RTL NN Accelerator. (Different Scales in the y-axis.)}}
\label{fig:vulRes}
\end{figure*}

\subsubsection{\textbf{When}, \textbf{\textit{Where}}, and \textbf{\textit{How frequently}} are Faults Injected?}
Our fault injection unit supports permanent (stuck-at-0 and stuck-at-1) and transient faults. Permanent faults can cause bits to permanently get stuck at 0 or 1, such as faults stemming from extremely low-voltage operation \cite{Chang2017}. Thus, they result in bit flips, if and only if logical values of faulty bits are different than the values at which the physical bits are stuck. In contrast, transient faults such as radiation induced faults \cite{radiation}, flip contents of bits for a few cycles (typically one cycle). Also, note that we evaluate the effect of both single-(single fault within a register) and multiple-bit (several faults within a register) faults to cover different fault rates. In short, we consider the following: 
\begin{itemize}
\item \textit{\textbf{When?}} Permanent faults are stuck at 0 or 1 for the whole inference cycles ($T$ as defined in Eq. \ref{T}), from first to the last. In contrast, transient faults occur in a single random cycle within inference $T$ cycles.
\item \textit{\textbf{Where?}} We inject faults into randomly-selected set of bits of a register of the RTL NN. 
\item \textit{\textbf{How frequently?}} To comprehensively study the impact of faults in the inference error, we repeat the fault injection for each input data item (for instance each image in MNIST). In other words, we generate a fault and while input data is streaming into the classifier one-by-one, we inject the generated fault for each of the input data items, individually. This accelerated fault injection campaign allows us to quickly evaluate the impact of faults on all input data items.
\end{itemize}

To generate faults, we use the pseudorandom number generator library of Bluespec, i.e., Randomize, to select a random register, random set of bits, and random cycle (in the transient case). Later on, we inject the generated fault in the corresponding locations/cycles, while streaming the input data into the classifier, allow the completion of the inference for all input data, and finally, compute the inference error by comparing classified against the golden output data provided by the benchmark suite.% We elaborate our statistical fault injection methodology in Section \ref{sec:statis}.%, where we experimentally determine the statistically significant number of trials.   

\subsubsection{Statistical Fault Injection Methodology} 
\label{sec:statis}
To achieve statistically significant results, we repeat experiments multiple times. In each trial, a different fault (a random register, set of bits, and cycle in the transient case) is randomly generated and injected. Finally, after repeating the injection for multiple times, the final inference error is calculated as the median of all these trials. If trials are repeated for significantly enough times, this statistical fault injection approach \cite{statistical} can lead to accurate results in comparison to the deterministic approach where all possible locations/cycles permutations are considered for the faults injection. Note that the deterministic approach is in practice hard to follow since there is billions of possibilities of faults generation.
%However, since the total number of possibilities for the fault injection is enormously large, in order of a billions with our RTL NN, it is almost impossible to examine all permutations in a deterministic approach. 
%
%The total number of bits, which can be potentially corrupted, can be computed using Equation \ref{S}, where $numR_i$ and $sizeR_i$ represent the number and size (bit-width) of register type $i \in \{IR, WR, IMR\}$, respectively.
%\begin{equation}
%\label{S}
%S = \sum_{i \in \{IR, WR, IMR\}}{numR_i * sizeR_i}
%\end{equation}  
%
%The number of possibilities for the fault injection is directly related to $S$ and $S * T$, for permanent and transient cases, respectively (As defined in Equations \ref{T} and \ref{S}, $T$ is the total number of classification cycles and $S$ is the total number of bits to be corrupted.). More specifically, for our DNN model with the baseline configuration introduced in Section \ref{sec1}, $S$= 4969 and $S * T$= 7890772. Due to large number of possibilities, it is almost impossible to compute the vulnerability of the design in a reasonable time by following the aforementioned deterministic solution. 
%Instead, we follow a statistical fault injection approach \cite{statistical}, where only a subset of possible faults is injected by repeating the trials. This subset is randomly selected with respect to the target injection bits and clock cycles. 
To find the enough number of trials note that it can be tuned according to the expected confidence level- the probability that the exact value is within a predefined error margin. % In this section, we experimentally extract the number of trials to achieve the statistical significant results with the confidence level= 90\% and error margin= 3\%. 
Experimental results for an example single-bit fault injection case are shown in Fig. \ref{fig:runs}, in terms of the aggregated median error rate of RTL NN in the y-axis and the number of trials in the x-axis (up to 1000). As can be seen, with error margin 1\%, 1000 trails leads to an acceptable confidence-level ($\sim$90\%) with a negligible standard deviation. Hence, we perform all our experiments 1000 times and use the median of these trials to report in this paper. %for instance, 421, 391, and 31 trials for stuck-at-1, stuck-at-0, and transient cases, are enough to achieve statistically acceptable results with confidence level= 90\% and error margin= 3\%. Note that this observed significant difference between the required number of trials for transient and permanent cases is the consequence of the momentary behavior of the transient faults and their subsequent relatively less impact on the NN error rate. In short, following this statistical analysis, we perform our experiments 1000 times and report their median in the rest of the paper.  %Note that as mentioned earlier, the total number of permutations to inject a transient fault is significantly more than the case of a permanent fault, 7890772 vs. 4969. However, we observe that to achieve a certain level of confidence, we need considerably less number of trials, in the transient against permanent case. We believe that is the consequence of the momentary nature of transient faults, where a transient bit corruption does not affect the overall DNN output considerably. 
\begin{figure}
\centering
\begin{subfigure}{0.24\textwidth}
\centering
\includegraphics[width=\linewidth]{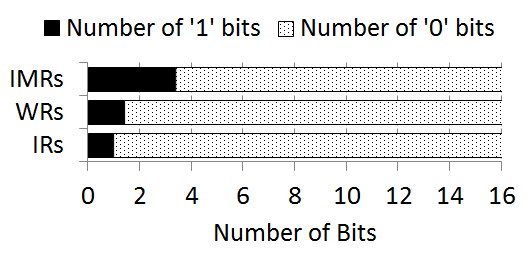} %, height=5cm
\caption{Statistical Sparsity Analysis of Inputs ($IR$s), Weights ($WR$s), and Intermediate ($IMR$s) Data.}
\label{fig:hist1}
\end{subfigure}
\hfill
\begin{subfigure}{0.24\textwidth}
\centering
\includegraphics[width=\linewidth]{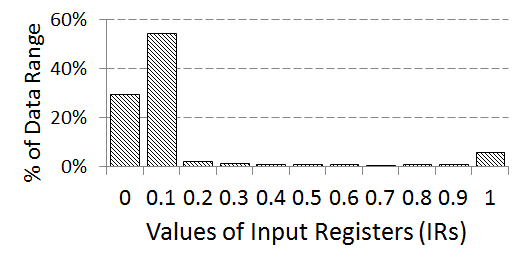} %, height=5cm
\caption{The Statistical Analysis in Ranges of Input Registers ($IR$s) Values.}
\label{fig:hist2}
\end{subfigure}
\caption{Histograms to show the sparsity of data of MNIST, used to justify; (a) why stuck-at-1 faults cause higher inference errors than stuck-at-0 faults, and (b) why \textit{IMR}s are more vulnerable than \textit{WR}s.}
\label{fig:hist}
\end{figure}

\section{Fault Characterization}
\label{sec:vul}
In this section, we evaluate the sensitivity of various components of our RTL NN against faults; i.e., application-level components (NN data, NN layers, and NN activation functions) and architectural-level components (data representation model and number of $PE$s). Toward this goal, we inject faults in these components, individually, by varying corresponding parameters of the baseline configuration as detailed in Table \ref{table:nnsetup}. Experimental results of the fault characterization are shown in Fig. \ref{fig:vulRes}. In this figure, the x-axis represents the number of injected faults per input object (image in MNIST), which varies from 0 for the fault-free NN case, to 16 for the case where all bits of the NN register are faulty. 

\subsection{Overall Effect of Different Fault Types in RTL NN}
We conduct the RTL NN fault characterization study in terms of the following fault categories:
\begin{itemize}
\item Permanent vs. Transient: As expected, permanent faults cause higher NN errors than transient faults. This is due to the persistence of permanent faults for the whole computation cycles ($T$ as explained in Eq. \ref{T}), while transient faults manifest themselves for a single cycle within $T$ cycles.   
\item Stuck-at-1 vs. Stuck-at-0: The permanent faults can be further categorized as being Stuck-at-1 or Stuck-at-0. As can be seen in Fig. \ref{fig:vulRes}, stuck-at-1 faults cause higher NN errors than stuck-at-0 faults. This observations is due to the sparsity of NN data, i.e., more '0' than '1' bits in the NN RTL registers, by on average 6.5X as can be seen in Fig. \ref{fig:hist1}.
\end{itemize}

\subsection{Application-Level Fault Analysis}
\subsubsection{NN Data}
The sensitivity of various NN data, i.e., inputs, weights, and intermediate data is evaluated by injecting faults in corresponding registers individually, i.e., $IR$s, $WR$s, and $IMR$s. Experimental results are shown in Fig. \ref{fig:103}, \ref{fig:104}, and \ref{fig:105}, for stuck-at-0, stuck-at-1, and transient cases, respectively. We observe that:
\begin{itemize}
\item $IR$s: Faults injection in $IR$s causes the relatively lowest inference error, since they are composed of only fraction component.
\item $WR$s and $IMR$s: In RTL NN, $IMR$s are the most vulnerable registers, due to two reasons; \textit{first}, as can be seen in Fig. \ref{fig:precision}, $IMR$s have relatively the longest digit component, which are significantly sensitive against faults. \textit{Second}, $IMR$s are used in the adder tree to maintain results of multipliers, as can be seen in Fig. \ref{fig:overall}. Therefore, any faults in $IMR$s is propagated to the next level of the adder tree without any coefficient. In contrast, $WR$s are inputs of multipliers and any fault in $WR$s is multiplied by $IR$s and then propagated. Nevertheless however, $IR$s accommodate very small values; as can be seen in Fig. \ref{fig:hist1}, 84.5\% of them are in the [0, 0.1] range, which limits the fault propagation in $WR$s.
%Thus, it can be concluded that \textit{WR}s (along with any fault) will be multiplied by a tiny values \textit{IR}s and then propagated, which leads to less error, in comparison to faults occurring in \textit{IMR}s. 
\end{itemize}

%In short, we observe that among NN data, intermediate ($IMR$s), weights ($WR$s), and inputs ($IR$s) in order have the highest vulnerability. 

\subsubsection{NN Layers}
The sensitivity of NN layers, i.e., $\{Layer_j, j \in [0,4]\}$ is evaluated by injecting faults in registers of these layers, individually. Experimental results are shown in Fig. \ref{fig:106}, \ref{fig:107}, and \ref{fig:108}, for stuck-at-0, stuck-at-1, and transient cases, respectively. We observe that:
%In this section, we study the behavior of faults in individual layers of the DNN. However, to perform a comprehensive study, we use a modified 5-layer ($\{L_i, i \in [0,4]\}$) DNN with the size of 784-512-256-128-10. Other baseline configurations of the RTL design are not changed, for instance the number of $PE$s= 64. Subsequently, with respect to Equation \ref{T}, the number of classification cycles can be computed as $T= 8852$, sum of $T_{layer0}=6272$, $T_{layer1}=2048$, $T_{layer2}=512$, and $T_{layer3}=20$, where $layer_j$ refers to the computation of $L_j$ and $L_{(j+1)}$, $j \in [0,3]$. Also note that for convenience, in this section, we inject faults within only \underline{non-sign bits} of \underline{\textit{{WR}}s} that leads to \textit{maxNumFaults= 15}. Furthermore, we modify the original fault injection unit, as elaborated below:
\begin{itemize}
\item Permanent Faults: Permanent faults in inner layers of the RTL NN, i.e., closer to the output layer, cause relatively lower inference error. Due to the persistence of permanent faults for the whole per-layer cycles, this observation is the consequence of relatively less cycles in inner layers, proportional to layer sizes, as detailed in Table \ref{table:nnsetup}.% Hence, since permanent faults are persistent for the whole per-layer cycles, they cause relatively less data corruption in relatively smaller inner layers.  
%less data corruption. By default, permanent faults stuck during the whole execution cycles. Instead, in this section, to infer the behavior of faults in the individual DNN layers, we stuck faults during the execution cycles of individual layers (not in the whole cycles). In other words, we stuck faults in 6272, 2048, 512, and 20 cycles of $layer_0$, $layer_1$, $layer_2$, $layer_3$ cycles, respectively (not in the whole $T$= 8852 cycles).  
\item Transient Faults: In contrast, transient faults in inner layers have relatively more impact on the inference error, since, \textit{first}, sizes of NN layers do not play any role for transient faults with momentary behavior, and \textit{second}, faults in inner layers have relatively less probability to be masked through the thresholding in activation functions of earlier layers.  
\end{itemize}

%With aforementioned points in the mind, we repeat experiments and show results in Figures \ref{fig:fault_layers_0}, \ref{fig:fault_layers_1}, and \ref{fig:fault_layers_2} for stuck-at-0, stuck-at-1, and transient cases, respectively. We observe that:
%\begin{itemize}
%\item \textbf{Permanent Faults}: Permanent faults in deeper layers of the RTL design of the DNN (closer to the output) cause lower classification errors. It is the consequence of smaller sizes of deeper layers in our modified DNN, which leads to significantly less execution cycles and subsequently, less data corruption (as computed earlier, 6272, 2048, 512, and 20 cycles for $layer_0$, $layer_1$, $layer_2$, and $layer_3$, respectively).
%\item \textbf{Transient Faults}: In contrast, we observe that deeper layers of the DNN are relatively more vulnerable, in the presence of transient faults. Considering the momentary nature of transient faults, we believe the reason is the impact of applying less number of activation functions in deeper layers, which prevent the fault propagation. In other words, in deeper layers any fault can be propagated to the final output of the DNN with less filtered by activation functions. 
%\end{itemize}   

%In short, we observe that in the presence of transient faults, inner layers are more vulnerable, in contrast to the permanent faults, where the vulnerability of NN layers is directly proportional to their sizes. 

\subsubsection{NN Activation Functions}
The activation function plays a crucial role in the overall accuracy and efficiency of NN. Hence, in this section, we study the fault propagation with two state-of-the-art NN activation functions, i.e., positive saturating linear (satlin) \cite{satlin} and logarithmic sigmoid (logsig) \cite{logsig} functions. %Although, many other activation functions exists, logsig and satlin are common and also only supported cases in MATLAB Autoencoders, which is used to achieve the significant training accuracy \cite{autoencoder}. 
Experimental results are shown in Fig. \ref{fig:112}, Fig. \ref{fig:113}, and Fig. \ref{fig:114}, for stuck-at-0, stuck-at-1, and transient cases, respectively. As can be seen, a relatively less error is observed by leveraging logsig against satlin. As already discussed, NN data are mainly near-zero. For this type of data, logsig applies a more aggressive quantification and thresholding than satlin, which in turn, limits the faults propagation.%leads to by on average 12.1\% relatively less NN classification error.% \raisebox{-\mydepth}{\fbox{\includegraphics[height=\myheight]{images/satlin.png}}} . 

%\begin{figure}
%\centering
%\begin{subfigure}{0.18\textwidth}
%\centering
%\includegraphics[width=\linewidth]{images/satlin.png} %, height=5cm
%\caption{Permanent Faults.}
%\label{fig:run_permanent}
%\end{subfigure}
%\hfill
%\begin{subfigure}{0.18\textwidth}
%\centering
%\includegraphics[width=\linewidth]{images/logsig.png} %, height=5cm
%\caption{Transient Faults.}
%\label{fig:run_transient}
%\end{subfigure}
%\caption{{\color{black}Repeating Trials to Achieve Statistically Significant Results. (Shows results of single-bit fault trials)}}
%\label{fig:runs}
%\end{figure}      

\subsection{Architectural-Level Fault Analysis}
\subsubsection{Different Components of Data Representation Model}
The sensitivity of various components of fixed-point data representation mode, i.e., sign, digit, and fraction is evaluated by injecting faults in corresponding components individually. Experimental results are shown in Fig. \ref{fig:100}, \ref{fig:101}, and \ref{fig:102}, for stuck-at-0, stuck-at-1, and transient cases, respectively. Note that for each component, the maximum number of injected faults is equal to the maximum bit-width of the corresponding components, as detailed in Fig. \ref{fig:precision}. We observe that sign, digit, and fraction components are in order, relatively more vulnerable, proportional to their positional significance, as expected.

\subsubsection{Parallelism Degree (Number of $PE$s)}
The number of $PE$s can impact the fault propagation behavior, since with respect to Eq. \ref{T}, it inversely affects the number of inference cycles ($T$). $T$ determines the data reuse degree, i.e., the number of register reloads to accomplish the inference, which can be directly translated to the persistence of permanent faults. To empirically perform this experiment, we repeat the fault injection for different number of $PE$s, i.e., 64, 256, and 1024. Fig. \ref{fig:109}, \ref{fig:110}, and \ref{fig:111} show experimental results in the presence of stuck-at-0, stuck-at-1, and transient faults, respectively. We observe that:
\begin{itemize}
\item Permanent Faults: In the presence of permanent faults, a relatively more $PE$s corresponds to proportionally reduced inference error. As said, it is the result of a relatively fewer number of clock cycles ($T$) and in turn, less persistence of permanent faults.
\item Transient Faults: In contrast, in the presence of transient faults, the number of $PE$s does not play any role, due to the momentary behavior of transient faults.
\end{itemize}

\subsection{Fault Characterization of Other NN Benchamrks}
We repeat the fault characterization study in Forest and Reuters benchmarks, as well, by following the similar experimental methodology explained in Section II. We observed that a vast majority of observations on MNIST, as discussed in Sections III-A and III-B, is staying valid for Forest and Reuters, as well. Experimental results of the fault propagation in those datasets are shown in Fig. \ref{fig:120}, \ref{fig:121}, and \ref{fig:122} for stuck-at-0, stuck-at-1, and transient cases, respectively. Due to our experimental observations, most important points are highlighted as follows:
\begin{itemize}
\item In our NN, the inherent fault-free inference error rate of MNIST, Forest, and Reuters are 2.56\%, 38.7\%, and 8.9\%, respectively, very close to the state-of-the-art implementation \cite{minerva}.
\item The rate of NN inference error increase of Reuter dataset in presence of additional permanent faults, is relatively more significant, since the size of the input layer of Reuter is considerable larger (2837 vs. 52/784), which in turn, leads to more persistence of permanent faults during the clock cycles ($T$). 
\end{itemize}

\section{Fault Mitigation}
\label{sec:mitigation}
In this section, we introduce and evaluate an efficient and low-overhead fault mitigation technique in RTL NN, relying on the fault characterization results. The aim is to recover from faults without leveraging any redundant bits as is common for traditional fault correction mechanisms such as Triple Modular Redundancy (TMR) \cite{tmr}. The proposed technique relies on the sparsity of NN data and accordingly, targets to mask faulty bits. This technique combines three individual mechanisms, i.e., \textit{first}, Word Masking to set all bits of the corrupted register to '0', \textit{second}, Bit Masking to mask faulty bit with the sign-bit within the faulty register, and \textit{third}, Sign-bit Masking to mask the sign-bit with the Most Significant Bit (MSB). Among them, Word- and Bit Masking techniques are proposed first time in Minerva \cite{minerva}; however, they are used individually and with a poor efficiency to protect the sign-bit, which in fact is relatively the most vulnerable component. To alleviate this issue, we propose to mask the sign-bit with the MSB, since through a statistical analysis we observed that with the probability of 99,9\% the sign-bit and MSB have the same logic value. This observation is the consequence of near-zero NN data and also two's complement data representation model, which finally leads to $sign\_bit= MSB$='0' for positive ($>0$) and $sign\_bit= MSB=$'1' for negative near-zero data ($<0$).% Also, we leverage these three methods in a unified way to take the advantage of them all.

%As comparison cases, we use fault mitigation techniques of a state-of-the-art DNN accelerator, Minerva \cite{reagen2016minerva}, since we similarly do not use redundant bits to recover bit flips. To have a fair comparison, we re-implement these techniques, including Word Masking (when a fault is detected, all bits of the corrupted register is set to zero) and Bit Masking (any bit that experiences the fault is replaced with the sign-bit). Both techniques rely on the sparsity of the MNIST; however, it is reported that Bit Masking technique is 44X more efficient. As noted earlier, the Bit Masking relies on the correctness of the sign-bit, since it is used as parity for other bits. However, as expected and we also observe in our experiments, a sign-bit corruption extirpates Bit Masking technique, since it does not propose any mechanism to handle the sign-bit flip. Our proposed techniques are aimed to alleviate this problem and mitigate faults, in which the sign-bit is corrupted. Also, we believe that our techniques do not incur much power, area, and delay overheads, since no redundant bits or additional circuits are used.      

\begin{figure}[!t]
\captionof{List}{Pseudo-code of Proposed Fault Mitigation Technique. }
\label{list:code}
\fbox{\begin{minipage}{0.49\textwidth}
%\centering
1: {if ({\color{black}\textbf{reg[N-1]}} \& {\color{black}\underline{reg[N-2]}} are flipped) reg $<=$ 0; \\
\hspace*{20ex}{{\textit{// WORD MASKING}}}}  \\  
2: {else begin}        \\
3: \hspace*{2ex}{if ({\color{black}\textbf{reg[N-1]}} {is} flipped) {\color{black}\textbf{reg[N-1]}} $<=$ {\color{black}\underline{reg[N-2]}};  \\
\hspace*{10ex}{{\textit{// SIGN-BIT MASKING WITH MSB}}}} \\
4: \hspace*{2ex}{for(i in [N-2, 0]) if (reg[i] {is} flipped) reg[i] $<=$ {\color{black}\textbf{reg[N-1]}}; \\
\hspace*{20ex}{{\textit{// BIT MASKING}} }} \\
5: {end}                              
\end{minipage}}
\end{figure}

It is important to note that about the fault detection we follow same assumptions with Minerva, i.e., there is no limit on the number of faults that can be detected and also, information is available on which bits are affected. To achieve them, in Minerva, Razor shadow latches \cite{Ernst2003} are simulated that can detect faults by monitoring circuit delays. %However, since we approach our study on the RTL design with an HLS approach, currently our design is not equipped with a fault detection technique. Instead, we detect faults by comparing original values of bits with their values after faults injection. If value of a bit after and before fault injection is not equal, it means a fault detection in the corresponding bit. 
ASIC \cite{minerva} and FPGA \cite{razorfpga} implementation of Razor report 0.3\% and 2.6\% of area overheads, respectively; which shows the feasibility of exploiting this method in the synthesize version of our RTL NN, as well.

The pseudo-code of the proposed fault mitigation technique is shown in List. \ref{list:code}. In this pseudo-code, the bit-width of the register (\textit{reg}) is assumed to be N, [N-1, 0], where {\color{black} \textbf{reg[N-1]}} and {\color{black} \underline{reg[N-2]}} are referring to the {\color{black} \textbf{sign-bit}} and {\color{black} \underline{MSB}}, respectively.
As can be seen, the proposed technique is composed of three sub-methods:
\begin{itemize}
\item Sign-Bit Masking with MSB: We use MSB as the mask of the sign-bit, since through an experimental analysis we observed these bits have same logic, with the probability of 99.9\%.
\item Bit Masking: Then, we apply Bit Masking technique to recover non sign-bit flips, by using sign-bit as the mask.
\item Word Masking: And finally, if MSB is itself also flipped, we reset the register to '0'. 
\end{itemize}

%figure
\begin{figure}
\centering
\begin{subfigure}{0.24\textwidth}
\centering
\includegraphics[width=\linewidth]{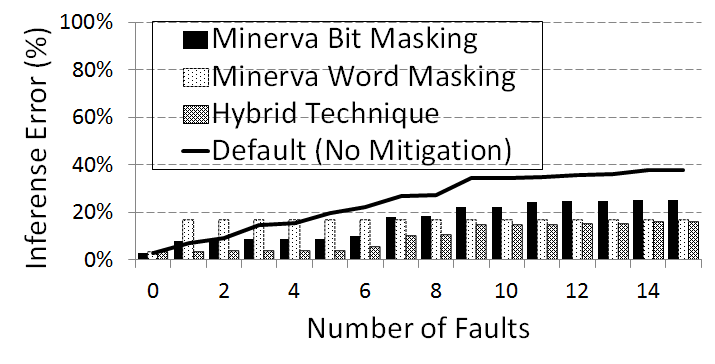} %, height=5cm
\caption{Evaluating Different Fault Mitigation Techniques (on MNIST).  }
\label{fig:mitigation2}
\end{subfigure}
\hfill
\begin{subfigure}{0.24\textwidth}
\centering
\includegraphics[width=\linewidth]{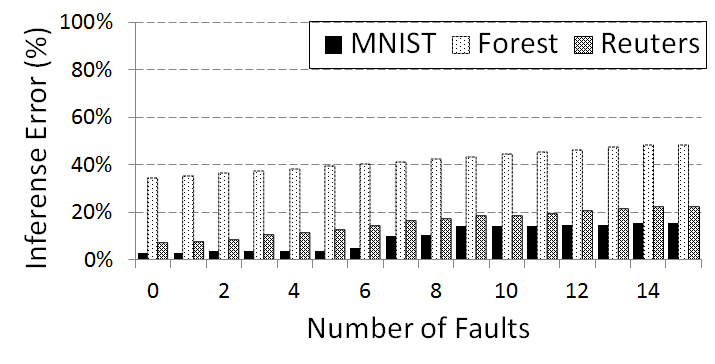} %, height=5cm
\caption{Enhanced Hybrid Technique on Different Datasets.}
\label{fig:mitigationdatasets}
\end{subfigure}
\caption{Evaluating Fault Mitigation Techniques (shown for stuck-at-0 case as similar efficiency observed for other types).}
\label{fig:mitigation}
\end{figure}

%\subsubsection{Evaluating}
We evaluate the proposed technique in our RTL NN and compare it with the individual Word- and Bit Masking techniques. Experimental results of the evaluation of these fault mitigation techniques are shown in Fig. \ref{fig:mitigation2}. As can be seen:
\begin{itemize}
\item Minerva Bit Masking: The efficiency of this technique is relatively the worst, since not only it does not have any mechanism to protect the sign-bit but also masks other faulty bits with sign-bit, which can be faulty. 
\item Minerva Word Masking: This technique leads to a constant inference error, independent of the number of detected faults, since the faulty register is reset to '0', when at least one fault is detected. 
%\item Word Flipping: This technique leads to a a constant 27.9\% classification error, since the whole register is flipped, in which at least one fault is detected in the corresponding register.
\item Proposed Enhanced Hybrid Technique: This technique shows relatively the best performance to correct faulty bits and achieve the lowest inference error, since it takes the advantage of both Bit- and Word Masking techniques and also, complement it by using MSB to mask the sign-bit faults.% We can infer form the curve that, \textbf{\textit{i)}} The sign-bit flip is always corrected by getting masked with MSB, since the error is not decreasing for additional faults. \textbf{\textit{ii)}} For less than (on average) 14 faults, the Bit Masking part of the technique is exploited, which effectively corrects faults by sign-bit masking. \textbf{\textit{iii)}} For more than (on average) 14 faults, the Word Masking part of this technique is leveraged (as both MSB and sign-bit are corrupted), which limits the error to 17.5\%. 
\end{itemize}

In short, our Enhanced Hybrid Technique mitigates faults and achieve by 47.3\% and 44.1\% lower NN inference error than Minerva Bit- and Word Masking techniques, respectively. 

%figure
%\begin{figure}
%\centering
%\begin{subfigure}{0.49\textwidth}
%\centering
%\includegraphics[width=\linewidth]{images/signbit.png} %, height=5cm
%\caption{Vulnerability analysis with the sign-bit corruption. }
%\label{fig:mitigation1}
%\end{subfigure}
%\begin{subfigure}{0.49\textwidth}
%\centering
%\includegraphics[width=\linewidth]{images/123.png} %, height=5cm
%\caption{Evaluating fault mitigation techniques.  }
%\label{fig:mitigation2}
%\end{subfigure}
%\caption{Evaluating Proposed Fault Mitigation Technique on Different Datasets. (shown for stuck-at-0 case as similar efficiency observed for other fault types).}
%\label{fig:mitigationdatasets}
%\end{figure}

\subsection{Fault Mitigation of Other NN Benchmarks}
We apply the proposed enhance hybrid fault mitigation technique on Forest and Reuters datasets, as well. The efficiency of the proposed fault mitigation method is shown in Fig. \ref{fig:mitigationdatasets}. By comparing these results against the default experiments without any protection in Fig. \ref{fig:120}, it can be observed that the proposed method has the capability to cover faults in all tested datasets, since the sparsity of data is an inherent feature of these benchmarks and it is also the base of the proposed technique. 

%%figure
%\begin{figure}
%\centering
%\fbox{\begin{minipage}{0.5\textwidth}
%\begin{enumerate}
%{\fontsize{7}{5}\selectfont \item  \textit{Bit\#(N) org\_reg; \textbf{// original corrupted register}}
%\item  \textit{Bit\#(N) corr\_reg = org\_reg; \textbf{// corrected register}} 
%\item \textit{if ({\color{red}org\_reg[N-1]} \& {\color{green}org\_reg[N-2]} \textbf{are} flipped) corr\_reg $<=$ 0; \\ \textbf{// Word Masking (similar to Minerva)}}
%\item   \textit{else begin}
%\item  \textit{if ({\color{red}org\_reg[N-1]} \textbf{is} flipped) {\color{red}corr\_reg[N-1]} $<=$ {\color{green}org\_reg[N-2]}; \\ \textbf{// MSB masking of the sign bit}}
%\item \textit{for(i in [N-2, 0]) if (org\_reg[i] \textbf{is} flipped) corr\_reg[i] $<=$ {\color{red}corr\_reg[N-1]};\\ \textbf{// Bit Masking (similar to Minerva)}}
%\item \textit{end}}
%\end{enumerate}
%\end{minipage}}
%\begin{subfigure}{0.44\textwidth}
%\centering
%\includegraphics[height=3.9cm]{images/mitigation.png} %, height=5cm
%%\label{fig:mitigationRes}
%\end{subfigure}
%\caption{Fault Mitigation: (left) Pseudo-code of Enhanced Hybrid Technique. (right) Experimental results.}
%\label{fig:mitigation}
%\end{figure} 

\section{related work}
\label{sec:related}
%our design both nn and fault injection unit are rtl and fully synthesizable. for instance below utilization on vc707. facilitate resilience study on the hardware XX XX
In this section, we review recent works on the resilience of NNs and highlight our contributions.
\subsection{Different Methodologies to Study NN Resilience}
It has been shown that NNs are inherently resilient \cite{dnnres1}, \cite{dnnres2}, \cite{dnnres3}; however, the ever-increasing fault rate in nano-scale technology nodes necessitates further studies in this area to explorer better trade-off of reliability, power, energy, and performance. Hence, in recent years NN resilience is studied with different approaches, e.g., software-level simulations or theoretical analyzes \cite{Tchernev}, \cite{Phatak}, SPICE simulations  \cite{minerva}, \cite{dnnfault}, \cite{ThUnderVolt}, and experimenting on the real hardware operating on low-voltage regimes \cite{dnnengine}, \cite{sram1}, \cite{sram2}. Among them, it is evident that software-level simulations and theoretical analyzes lack the information of the underlying hardware platform and are relatively less precise. In contrast, SPICE-based studies are more precise; however, these studies require significant circuit-level efforts. Our aim is to have the best of both worlds: to have the flexibility and simplicity of  software-level fault-injection with the precision of circuit-level implementations. Thanks to the evolution of HLS tools, the precise modeling of the underlying hardware has been facilitated, which provides an opportunity to perform such studies comprehensively with an accuracy close to the real hardware. This approach is also followed by some recent works to study the resilience of several other applications \cite{tambara2017}, \cite{lee2017}, \cite{tonfat2017}, \cite{dossantos2017}.    

\subsection{Fault Characterization on NNs}
The impact of faults in different models of NNs have been studied in a wide body of recent work as surveyed in \cite{dnnres1}. Among the most relevant existing works, Minerva \cite{minerva} performs a characterization on the sparsity of data and analyzes the efficiency of leveraging fixed-point data representation model. In the same line, \cite{dnnres3} studied the vulnerability of various layers of NN. Also, recently \cite{dnnfault} studied the fault propagation in an ASIC model of NN focused on the vulnerability of different NN layers. However, these works does not present comprehensive characterization studies, whereas our paper comprehensively studies the sensitivity analysis of both application- (NN data, NN layers, and activation function) and architectural-level (parallelism degree and data representation model) components. This comprehensiveness is the consequence of using HLS tools to facilitate the RTL modeling of the NNs. 
   
\subsection{Fault Mitigation on NNs}
To mitigate faults several general techniques are proposed in different domains such as TMR \cite{tmr}, Razor \cite{Ernst2003}, \cite{Stott2014}, ECC using Hamming code \cite{ecc}, Hardware Transnational Memory (HTM) \cite{htm}, among others. These techniques can be potentially customized to mitigate faults of NNs, as well; however, with timing, area, or power costs. Also, techniques adapted for NNs are surveyed in \cite{Torres}, such as explicit redundancy, retraining, and modifying learning/inference phases. In this paper, instead of costly fault tolerance operations we present an application-aware fault mitigation in NNs, which does not require to exploit any redundant data bits or additional considerable overheads. In the same line, Minerva \cite{minerva} has also proposed two fault mitigation techniques, i.e., bit masking and word masking, which rely on the sparsity of NN data. In our paper, we use these techniques as baseline comparison cases. We explored their efficiency issues when the sign-bit is corrupted and accordingly, presented a more efficient technique to be effective in such cases, as well. %However, the Minerva work lacks of an in-depth fault characterization study. 
%In this paper, we thoroughly study the behavior of faults in the RTL design of a DNN model and propose appropriate fault mitigation techniques. 
%Authors in \cite{yang2017sram} studied the impact of reduced-voltage Static RAM (SRAM) in the resilience of CNN. However, the fault characterization study is not comprehensive, as only input data are located in SRAMs, not weights for example. Also, exploiting SRAMs to reside large volumes of input data is not efficient, considering their costs and low densities. Its other drawback is that the faulty SRAM data are exported into a software CNN to study the accuracy. Means that CNN is assumed to be fault-free, which is not essentially true. 

%mention to real DNN models, their problem is that they need much efforts

\section{conclusion and future work}
\label{sec:conc}
We empirically investigated an RTL hardware-aware design of NN accelerator from the resilience perspective, i.e., fault characterization and mitigation. We comprehensively characterized the vulnerability of various components of the design, in the presence of permanent and transient faults. Our experiments indicate that the following parameters play crucial roles in the fault severity: data representation mode, NN data (inputs, weights, or intermediate), NN layers, activation function, and parallelism degree of the underlying accelerator. Relying on the characterization results, we proposed a low-overhead fault mitigation technique to efficiently correct corrupted bits of RTL NN. Our proposed technique can mitigate faults, by 47.3\% better than state-of-the-art methods. 

We plan to perform fault characterization and mitigation study on more advanced RTL NN models, e.g., Convolutional NNs (CNNs) and Recursive NNs (RNNs), as well.
% no keywords

% For peer review papers, you can put extra information on the cover
% page as needed:
% \ifCLASSOPTIONpeerreview
% \begin{center} \bfseries EDICS Category: 3-BBND \end{center}
% \fi
%
% For peerreview papers, this IEEEtran command inserts a page break and
% creates the second title. It will be ignored for other modes.
\IEEEpeerreviewmaketitle

\section*{Acknowledgment}
We thank Pradip Bose, Alper Buyuktosunoglu, and Augusto Vega from IBM Watson for their contribution to this work. The research leading to these results has received funding from the European Union's Horizon 2020 Programme under the LEGaTO Project (www.legato-project.eu), grant agreement n$^\circ$ 780681.

% that's all folks
\end{document}